\documentclass[11pt]{article}

\usepackage[preprint]{acl}

\usepackage{times}
\usepackage{latexsym}
\usepackage[T1]{fontenc}
\usepackage[utf8]{inputenc}

\usepackage{microtype}
\usepackage{inconsolata}
\usepackage{booktabs} 
\usepackage{tabularx}

\usepackage{color}
\usepackage{colortbl} 
\usepackage{subcaption}

\usepackage{xcolor}
\usepackage{graphicx}

\usepackage[cjk]{kotex}
\newcommand{\zh}[1]{\begin{CJK}{UTF8}{gbsn}#1\end{CJK}}
\usepackage{amsmath,amssymb}
\usepackage{amsthm}

\usepackage{algorithmic,algorithm}
\renewcommand{\algorithmiccomment}[1]{\bgroup\hfill$\triangleright$~#1\egroup}

\newtheorem{lemma}{Lemma}

\newenvironment{proofsketch}{\par\noindent\textit{Proof sketch}\ }{\hfill$\square$\par}

\usepackage{langsci-gb4e}

\title{Chinese Word Boundary Recovery through Character Alignment Projection}

\author{
Lusha Wang\raisebox{-0.1\height}{\includegraphics[height=0.7em]{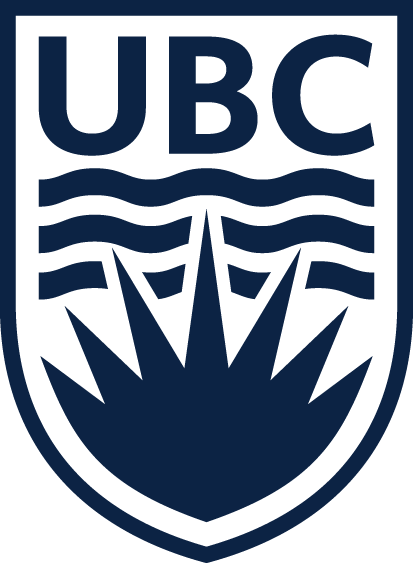}}$^{\dagger}$ \quad Yuchen Li\raisebox{-0.1\height}{\includegraphics[height=0.7em]{UBC.png}}$^{\dagger}$ \quad Su Yuan\raisebox{-0.1\height}{\includegraphics[height=0.7em]{UBC.png}}$^{\dagger}$ \quad 
Jungyeul Park\raisebox{-0.1\height}{\includegraphics[height=0.7em]{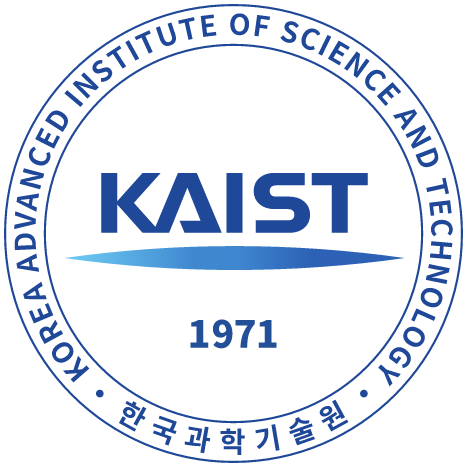}}\thanks{Corresponding author; $\dagger$Equally contributed authors.} \\
\raisebox{-0.1\height}{\includegraphics[height=0.9em]{UBC.png}} The University of British Columbia, Vancouver, Canada\\
\raisebox{-0.1\height}{\includegraphics[height=0.9em]{KAIST}} Korea Advanced Institute of Science \& Technology, Daejeon, South Korea\\
{\tt \{lushaw12,yyyuchen,yuansu\}@student.ubc.ca} \quad {\tt jungyeul@kaist.ac.kr}
\\}

\begin{document}
\maketitle
\begin{abstract}
Chinese word segmentation is especially fragile in non-standard text, where language learner errors and other character-level divergences disrupt the word boundaries assumed by downstream annotation and evaluation. This paper formulates Chinese word boundary recovery as an alignment-based projection task. Given a noisy source sentence and a cleaner target counterpart, we first align the two strings at the character level and then project target-side word boundaries back onto the source. Beyond the recovery method itself, we introduce two evaluation resources: a manually checked learner Chinese benchmark based on MuCGEC and a controlled synthetic benchmark derived from the Chinese Penn Treebank. 
Experiments show that direct segmentation remains vulnerable to compound fragmentation in learner input, whereas the proposed two step projection method corrects many over-segmentation errors by using the corrected target to recover source-side word spans. The results show that word boundary recovery is distinct from ordinary segmentation and that alignment projection provides a principled mechanism for stabilizing Chinese annotation and evaluation under noisy input.
\end{abstract}

\section{Introduction}

Chinese natural language processing begins with the deceptively simple task of detecting word boundaries.
Unlike alphabetic languages such as English, written Chinese is a continuous string of characters without explicit delimiters.
Segmentation is therefore a prerequisite for many downstream tasks, from part-of-speech tagging to parsing \citep{sproat-etal-1996-stochastic,sproat-emerson-2003-first}.
Errors at this stage are not merely local.
A single mis-segmentation can propagate upward, affecting syntactic analyses, edit extraction, evaluation metrics, and even the feedback presented to human learners.

This problem becomes more acute in learner Chinese \citep{linlin-2006-error,dazhong-2020-analysis}.
Learner sentences often contain character substitutions, omissions, additions, and local word-order deviations, all of which may disrupt the word boundaries assumed by standard segmentation systems.
For example, if a learner produces \zh{旅行困} \textit{l\v{\"u}x\'ing kùn} (`travel + tired') instead of the intended \zh{旅行团} \textit{l\v{\"u}x\'ingtu\'an} (`tour group'), a segmentation system may split the learner form as \zh{旅行$\sqcup$困}, where $\sqcup$ denotes an implicit space, while treating the corrected form as a single word.
In Chinese grammatical error correction, such a local mismatch has broader consequences: edits may be aligned to the wrong units, corrections may be misclassified, and feedback may be attached to distorted word spans rather than to the intended expression.
Word boundary errors therefore compromise not only tokenization, but also annotation quality, evaluation reliability, and interpretability.

A direct segmentation approach treats the learner source in isolation.
This is insufficient when the goal is to recover the intended word units of a noisy sentence.
Learner data often comes with a cleaner counterpart, namely a manually corrected target sentence.
The central question is therefore not simply how to segment the source string, but how to recover source-side word boundaries with respect to the corrected target.
Chinese characters provide a natural substrate for this problem: they are smaller than word-level segmentation units, but stable enough to support alignment across source and target variants.
By aligning source and target characters, word boundaries from the corrected target can be projected back onto the learner source.
This reframes Chinese word boundary recovery as a monolingual projection problem, conceptually related to annotation projection \citep{yarowsky-ngai-2001-inducing,hwa-EtAl:2005}, but focused on recovering internal segmentation structure under learner-induced character divergence.

This paper proposes a two step character alignment projection framework for Chinese word boundary recovery.
The first step, denoted \(\mathcal{P}_1\), constructs a monotone character alignment and retains identical character matches as high-confidence anchors.
These anchors provide reliable fixed points for projection and make it possible to localize the regions where the learner source and corrected target diverge.
The second step, denoted \(\mathcal{P}_2\), aligns the remaining unaligned characters using complementary similarity signals, including glyph shape, phonological similarity, relative position, and contextual semantic similarity.
For each alignment setting, the resulting character correspondences are passed to the same deterministic projection operator, which transfers target-side word boundaries to the source while preserving source coverage and contiguous token spans.
The framework therefore improves recovery by enriching the alignment in two stages: first with reliable identical-character correspondences, and then with similarity-based links where exact matching is insufficient.

This paper makes three contributions.
First, we formulate Chinese word boundary recovery as an alignment-based projection task, in which the segmentation of a corrected target sentence is transferred to its noisy learner source counterpart.
Second, we propose a two step character alignment method that combines identical-character anchoring with similarity-based alignment over remaining unaligned characters, together with a deterministic projection operator for source-side word boundary recovery.
Third, we construct two evaluation resources for this task: a manually checked learner Chinese benchmark based on MuCGEC \citep{zhang-etal-2022-mucgec} and a controlled synthetic benchmark derived from the Chinese Penn Treebank \citep{xue-etal-2005-ctb}, enabling evaluation under both naturally occurring learner errors and systematically varied character-level noise.

\section{Method from alignment to projection}
\label{sec:method}

We cast Chinese word boundary recovery as a target-conditioned projection problem over paired learner sentences.
Given a noisy learner source string \(S\) and a corrected target sentence \(T\) with word boundaries \(W_T\), the goal is to recover a source-side tokenization \(\widehat{W}_S\) that preserves the original learner string while reflecting the lexical boundary structure licensed by the corrected target.
The task therefore differs from ordinary Chinese word segmentation: the source is not segmented in isolation, but with respect to a cleaner counterpart that provides reliable word boundary evidence.

The method has two components.
First, a character alignment module constructs correspondences between \(S\) and \(T\).
Second, a deterministic projection operator transfers target-side word boundaries through these character correspondences and emits a recovered source tokenization.
This separation makes the role of alignment explicit: different alignment assumptions can be supplied to the same projection operator, while the projection procedure itself remains fixed.

\subsection{Task formulation}
\label{subsec:setup}

Let \(S=s_0\cdots s_{m-1}\) be a noisy learner source string, and let \(T=t_0\cdots t_{n-1}\) be its corrected target counterpart.
The source may come with an initial tokenization \(W_S^0\), while the corrected target is segmented as \(W_T\).
In this paper, \(W_T\) follows a word-oriented, compound-based segmentation scheme consistent with the Chinese Penn Treebank \citep{xue-etal-2005-ctb}.
The recovered tokenization \(\widehat{W}_S\) is evaluated against source-side gold spans under the same granularity.

We assume a character-level alignment relation \(\mathcal{A}\subseteq \{0,\ldots,m-1\}\times\{0,\ldots,n-1\}\).
A link \((i,j)\in\mathcal{A}\) indicates that source character \(s_i\) corresponds to target character \(t_j\).
Source or target characters without links represent deletion, insertion, or otherwise unresolved material.
Given \(S\), \(W_S^0\), \(T\), \(W_T\), and \(\mathcal{A}\), the projection operator computes the recovered source tokenization \(\widehat{W}_S\).

\subsection{Two step character alignment}
\label{subsec:two-step-alignment}

The proposed framework constructs character correspondences in two steps.
The first step, denoted \(\mathcal{P}_1\), uses identical-character alignment.
A Levenshtein-style dynamic program produces an order-preserving edit path between \(S\) and \(T\), but only identical-character matches are retained as reliable links.
These links form \(\mathcal{A}_1\), a high-precision alignment backbone that anchors the shared material between the learner source and the corrected target.
Projection through \(\mathcal{A}_1\) tests how much word boundary recovery can be achieved from exact character identity alone.

The second step, denoted \(\mathcal{P}_2\), extends this backbone by aligning the remaining unaligned characters.
Let \(U_S\) and \(U_T\) be the source and target indices not covered by \(\mathcal{A}_1\).
For each candidate pair \((s_i,t_j)\), where \(i\in U_S\) and \(j\in U_T\), we compute a weighted similarity score combining glyph similarity, phonological similarity, relative position, and contextual semantic similarity.
High-scoring pairs are selected under a one-to-one constraint, yielding a residual alignment \(\mathcal{A}_2\).
The final alignment used by \(\mathcal{P}_2\) is
$\mathcal{A}
=
\mathcal{A}_1 \cup \mathcal{A}_2 \cup \mathcal{A}_{\varnothing}$,
where \(\mathcal{A}_{\varnothing}\) records unresolved source or target characters.

This design uses exact identity where it is maximally reliable and applies similarity-based evidence only to residual regions where the source and target diverge.
The two steps therefore correspond to two progressively richer projection settings:
\(\mathcal{P}_1\) projects word boundaries through identical-character anchors, while \(\mathcal{P}_2\) projects through both identical-character anchors and similarity-based residual links.
Detailed feature definitions and scoring are given in Appendix~\ref{app:similarity-features}.

\subsection{Boundary projection}
\label{subsec:projection}

Given a fixed alignment \(\mathcal{A}\), the projection operator
\[
\Pi(S,W_S^{0},W_T,\mathcal{A})=\widehat{W}_S
\]
transfers target-side word boundary evidence to the source.
The operator represents the source segmentation as a set \(B_S\) of word-final source character positions, initialized from the initial source tokenization \(W_S^0\).
It then updates \(B_S\) in two conservative ways.

First, a target-side boundary is inserted into the source only when the two adjacent target characters align to adjacent source characters.
This prevents boundaries from being projected across insertion, deletion, or non-local alignment regions.
Second, a source-internal boundary is removed only when a target token projects to a complete contiguous source span.
This corrects source-side oversegmentation when a target word licenses a corresponding source span.
All ambiguous regions retain the initial source segmentation.

\begin{algorithm}[!t]
\footnotesize
\caption{\textsc{Word boundary projection}}
\label{alg:projection}
\begin{algorithmic}[1]
\STATE \textbf{Input:} source sentence \(S\), initial source segmentation \(W_S^{0}\), target segmentation \(W_T\), character alignment \(\mathcal{A}\)
\STATE \textbf{Output:} projected source segmentation \(\widehat{W}_S\)
\STATE Initialize source boundary set \(B_S\) from \(W_S^{0}\)
\STATE Extract target boundary set \(B_T\) from \(W_T\)
\FOR{each target boundary \(b\in B_T\)}
    \IF{the adjacent target characters align to adjacent source characters}
        \STATE Insert the corresponding source boundary into \(B_S\)
    \ENDIF
\ENDFOR
\FOR{each target token \(w_T\in W_T\)}
    \IF{\(w_T\) projects to a complete contiguous source span}
        \STATE Remove source-internal boundaries inside that span
    \ENDIF
\ENDFOR
\STATE Ensure that the final source character is marked as a boundary
\STATE \(\widehat{W}_S\gets\) segmentation of \(S\) induced by \(B_S\)
\STATE \textbf{return} \(\widehat{W}_S\)
\end{algorithmic}
\end{algorithm}

For a fixed alignment, \(\Pi\) is deterministic.
It preserves full source coverage, emits only contiguous source spans, and does not alter the learner string itself.
Thus, the method recovers word boundaries rather than correcting the source sentence.
Formal properties and complexity of the projection operator are given in Appendix~\ref{app:projection-properties}.

\section{Evaluation benchmarks}
\label{sec:benchmarks}

We construct two benchmarks for Chinese word boundary recovery.
The first is a manually checked learner Chinese benchmark based on MuCGEC \citep{zhang-etal-2022-mucgec}, which evaluates projection under naturally occurring learner errors.
The second is a synthetic benchmark derived from the Chinese Penn Treebank 5.1 \citep{xue-etal-2005-ctb}, which evaluates robustness under controlled character-level noise.
Both benchmarks use the same word-oriented, compound-based segmentation granularity, following the conventions of CTB and LTP \citep{che-etal-2021-n}.
Additional construction details and statistics are provided in Appendix~\ref{app:benchmark-details}.

\subsection{Learner Chinese benchmark}
\label{subsec:benchmark-learner}

The learner benchmark is constructed from the development portion of MuCGEC.
Each instance consists of a learner source sentence and one or more human-corrected target sentences.
Because the proposed projection framework requires a single corrected target from which word boundaries can be projected, we first select one target correction for each learner source.

\paragraph{Reference selection}
MuCGEC provides multiple human corrections for many learner sentences.
We select a single corrected target using a two-stage procedure.
First, candidate corrections are ranked using two automatic criteria: cosine similarity over pretrained character-level sentence representations and Levenshtein distance between the learner source and each corrected target.
When the two criteria select the same correction, that correction is retained automatically.
This procedure selects 765 source--target pairs without manual intervention.

The remaining 314 cases are manually adjudicated.
Three annotators independently choose the most appropriate correction according to two criteria: grammatical well-formedness of the corrected sentence and semantic fidelity to the learner source.
Final targets are determined by majority vote.
Among the manually adjudicated cases, 115 sentences receive full agreement among the three annotators, while 199 sentences receive a two-of-three majority decision.
No sentence remains disputed after majority voting.
The moderate agreement at this stage reflects the nature of MuCGEC, where multiple corrections may be grammatical and semantically plausible.
For word boundary recovery, however, majority voting provides a deterministic target while preserving the intended meaning of the learner source.

\paragraph{Segmentation granularity}
Chinese word segmentation is not uniquely defined, and different resources may follow different conventions.
For this benchmark, we adopt a word-oriented, compound-based segmentation scheme consistent with CTB.
This choice is motivated by the goal of recovering lexical word boundaries rather than morpheme-level internal structure.

Under this scheme, stable multi-character lexical units are treated as single tokens when they function as lexicalized words.
For example, \zh{旅行团} \textit{l\v{\"u}x\'ingtu\'an} (`tour group') is treated as one token rather than as \zh{旅行} \textit{l\v{\"u}x\'ing} (`travel') plus \zh{团} \textit{tuán} (`group').
Similarly, \zh{现代化} \textit{xiàndàihuà} (`modernize') is treated as a compound-level token when it functions as a lexicalized unit.
The same segmentation granularity is used for both the learner benchmark and the synthetic CTB-derived benchmark, ensuring that results are directly comparable across the two evaluation settings.

\paragraph{Gold source-side annotation}
After target selection, all learner sources are manually segmented under the compound-based guideline.
The annotation goal is to produce a source-side gold tokenization that covers the original learner sentence while reflecting the intended lexical units recoverable from the selected corrected target.
This is necessary because learner Chinese may contain character substitutions, omissions, insertions, or local word-order deviations, all of which can disrupt word boundaries.

The annotation process begins with system-generated boundary proposals.
Three annotators independently correct these proposals according to the compound-based guideline and established Chinese segmentation criteria \citep{xia-2000-segmentation}.
Residual disagreements are adjudicated to produce the final gold source segmentation.
Annotator consistency is high, with boundary-level pairwise \(F_1=0.987\) and Krippendorff's \(\alpha=0.981\).
The final learner benchmark contains 1,079 source--target pairs.

\subsection{Synthetic noise benchmark}
\label{subsec:benchmark-synthetic}

The learner benchmark reflects naturally occurring learner errors, but it does not control the amount or type of source--target divergence.
We therefore construct a synthetic benchmark from CTB 5.1.
Clean CTB segmented sentences are used as targets, and noisy source sentences are generated by injecting character-level substitutions, deletions, and insertions.
The purpose of this benchmark is not to simulate learner Chinese exhaustively, but to provide controlled test conditions under which character-level divergence can be increased systematically.

\paragraph{Clean target data}
The clean target side is obtained from segmented CTB 5.1 sentences.
Because CTB already provides word-oriented segmentation, these clean sentences serve as the gold target tokenizations.
We select a fixed subset of CTB sentences, IDs 400--550, and use the same base subset across all noise levels.
This ensures that differences across synthetic conditions are attributable to the amount of injected character-level noise rather than to changes in sentence selection.

\paragraph{Noise distribution}
The relative proportions of synthetic edits are estimated from the learner benchmark.
For each learner source and selected corrected target pair, we compute a character-level sequence alignment and map the resulting edit operations to three learner-perspective error types: substitution, deletion, and insertion.
Aggregating these operations yields the following distribution:
$p(\textsc{sub}) = 0.483$,
$p(\textsc{del}) = 0.318$, and
$p(\textsc{ins}) = 0.199$.
This distribution controls only the relative frequency of synthetic edit operations.
It does not determine the segmentation granularity, which remains inherited from the clean CTB reference.

\paragraph{Noise injection}
For each synthetic condition, we specify a global noise ratio \(r\), where \(r\in\{1\%,2\%,\ldots,10\%\}\).
The ratio is defined as the number of character-level edit operations divided by the total number of characters in the selected CTB subset.
Noise is injected globally across the sampled sentences.
Each edit operation affects one character position and is sampled as a substitution, deletion, or insertion according to the learner-derived distribution above.
Edit locations are sampled at the character level and are independent of gold word boundaries.
For substitutions, a source character is replaced with a character sampled from a global character pool.
For deletions, a character is removed from the noisy source sequence.
For insertions, an additional character sampled from the global character pool is inserted into the noisy source sequence.
The clean CTB sentence remains the target, while the perturbed sentence becomes the noisy source.

\paragraph{Synthetic source--target pairs}
Each synthetic example pairs a noisy source sentence with its corresponding clean segmented CTB target sentence.
The clean target provides the reference word boundaries, while the noisy source contains the injected character-level perturbations.
Inserted source characters have no corresponding characters in the clean target, and deleted target characters are absent from the noisy source sequence.
The same CTB sentence subset is used at every noise level, with a fixed but distinct random seed for each noise ratio.
The resulting benchmark contains ten evaluation conditions, one for each noise ratio from \(1\%\) to \(10\%\), allowing direct segmentation and alignment-based projection methods to be evaluated as source--target divergence increases under controlled perturbation.

\section{Experiments and results}
\label{sec:experiments}

We evaluate Chinese word boundary recovery on two evaluation benchmarks. The experiments address three questions.
First, does target-conditioned projection improve over direct segmentation of the noisy source?
Second, does the full two step projection method improve over projection based only on identical-character anchors?
Third, how robust are the systems under controlled increases in character-level synthetic noise?

\subsection{Systems and evaluation protocol}
\label{subsec:systems}

We compare three systems: direct segmentation, \(\mathcal{P}_{1}\) (identical-character projection), and \(\mathcal{P}_{2}\) (similarity-based projection).

\paragraph{Direct segmentation}
The direct baseline, denoted \(\mathcal{D}\), applies the automatic word segmentation tool \citep{che-etal-2010-ltp} directly to the noisy source sentence \(S\).
This represents the standard setting in which learner input is segmented without access to its corrected target.

\paragraph{\(\mathcal{P}_{1}\)}
The first projection system uses only identical-character anchoring.
It identifies an order-preserving set of identical character correspondences between the source and target strings and retains these matches as reliable links.
Unlike a full edit-distance alignment, this step does not resolve substitutions: non-identical characters are left unresolved and are handled conservatively by the projection operator \(\Pi\).
This system tests how much word boundary recovery can be achieved from exact character anchoring alone.

\paragraph{\(\mathcal{P}_{2}\)}
The full projection system extends \(\mathcal{P}_{1}\) by aligning remaining unaligned characters using similarity-based evidence.
The residual alignment uses glyph similarity, phonological similarity, relative position, and contextual semantic similarity.
The final alignment combines identical-character links, similarity-based residual links, and unresolved null material.
The same projection operator \(\Pi\) is then used to produce the recovered source tokenization.

\paragraph{Evaluation protocol}
\label{subsec:evaluation-protocol}

All systems are evaluated as word boundary recovery systems over the noisy source sentence.
For each source sentence \(S\), the system outputs a recovered tokenization \(\widehat{W}_S\), which is compared against the gold source-side tokenization.
We report micro-averaged token-level \(F_1\).\footnote{A predicted token is counted as correct only when its character span exactly matches a gold token span.
This metric directly reflects the goal of word boundary recovery: the system must reconstruct complete word units on the noisy source side, rather than merely place correct boundary offsets.
Metrics are computed with \textsc{Jp-preprocessing} \citep{jo-etal-2024-untold-story}.
}
For projection systems, the target sentence \(T\) is segmented first with the automatic word segmentation tool, and its word boundaries are projected to \(S\) through the corresponding character alignment.
Thus, \(\mathcal{P}_{1}\) and \(\mathcal{P}_{2}\) differ only in the alignment \(\mathcal{A}\) supplied to the same projection operator \(\Pi\).

\subsection{Main results on learner Chinese}
\label{subsec:main-learner-results}

Table~\ref{tab:learner-main-results} reports token-level word boundary recovery results on the learner Chinese benchmark.
Direct segmentation provides a strong baseline, but it segments the learner source in isolation and cannot use boundary evidence from the corrected target.
Projection with identical-character links improves over direct segmentation, showing that reliable character correspondences already provide useful target-side boundary information.
The full two step projection achieves the best performance by additionally resolving non-identical but related source--target characters.

\begin{table}[!ht]
\centering
\footnotesize
\resizebox{.48\textwidth}{!}{
\begin{tabular}{lcc}
\toprule
Method & Token \(F_1\) & \(\Delta\) prev. \\
\midrule
\(\mathcal{D}\) \ Direct segmentation
& 0.9891 & -- \\
\(\mathcal{P}_{1}\) Identical-character projection
& 0.9899 & +0.0008 \\
\(\mathcal{P}_{2}\) Similarity-based projection
& 0.9916 & +0.0017 \\
\bottomrule
\end{tabular}
}
\caption{Token-level word boundary recovery results on the learner Chinese benchmark. 
}
\label{tab:learner-main-results}
\end{table}

A paired significance test confirms that \(\mathcal{P}_{2}\) improves reliably over both \(\mathcal{P}_{1}\)
(\(p = 0.0006\)) and \(\mathcal{D}\) (\(p = 0.0026\)), while \(\mathcal{D}\) and \(\mathcal{P}_{1}\) do not differ
significantly (\(p = 0.2690\)). Although the absolute gain is modest, the
learner benchmark is a high-baseline setting: most source-side word boundaries
are already recoverable by direct segmentation. The improvement of \(\mathcal{P}_{2}\) is
therefore concentrate \(\mathcal{D}\) in the residual cases where learner-side character
divergence disrupts lexical word recognition. The non-significant gain from  \(\mathcal{D}\) 
to \(\mathcal{P}_{1}\) shows that exact character anchoring alone provides limited benefit, while
the reliable gain from \(\mathcal{P}_{1}\) to \(\mathcal{P}_{2}\) shows that similarity-based residual alignment
contributes beyond exact matching. This effect is further supported by the error
analysis, where \(\mathcal{P}_{2}\) reduces over-segmentation, and by the synthetic benchmark,
where \(\mathcal{P}_{2}\) remains more stable as character-level noise increases.

\subsection{Results on synthetic noise}
\label{subsec:synthetic-results}

\begin{figure}[!ht]
\footnotesize
\centering
\includegraphics[width=0.9\linewidth]{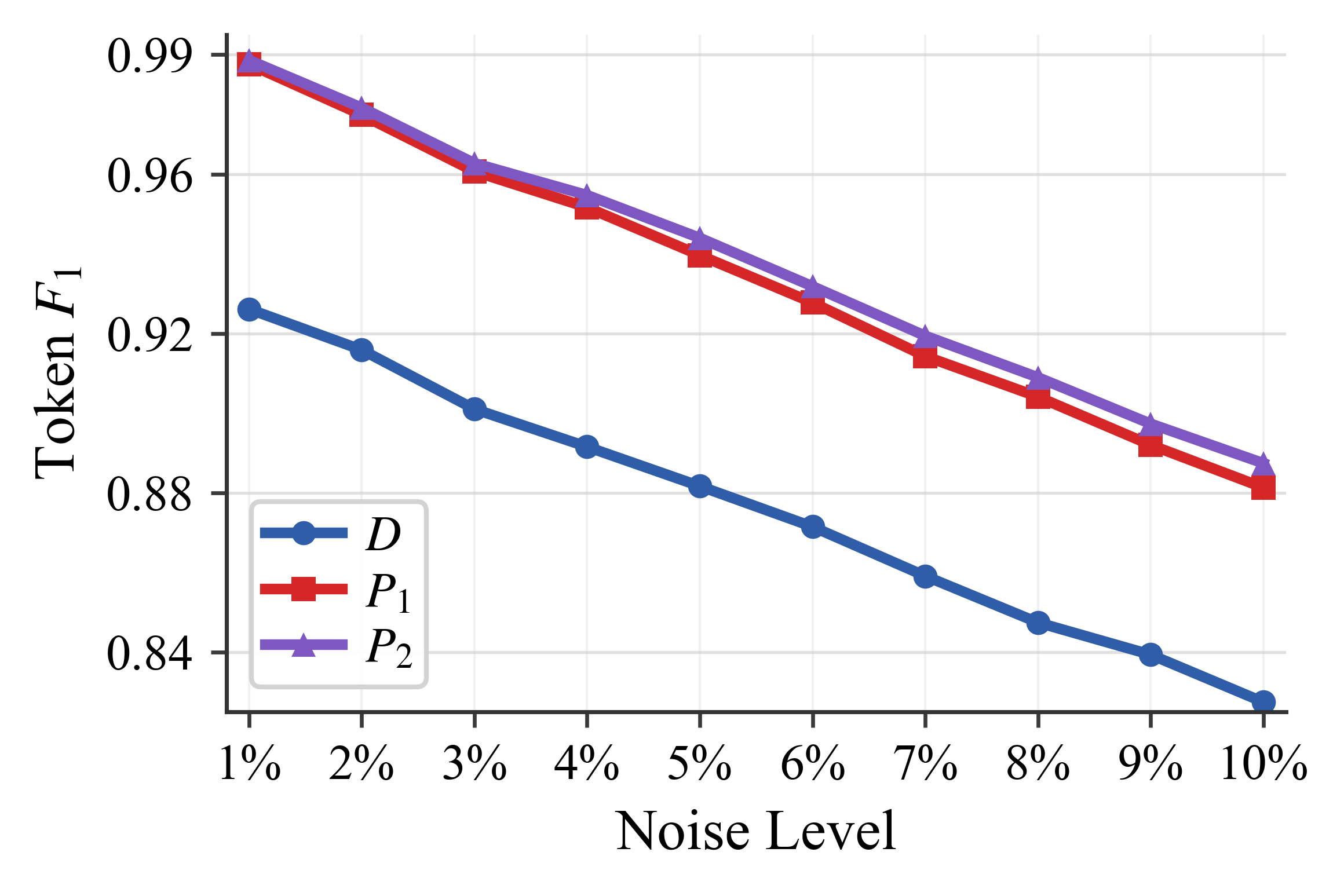}
\caption{Token-level \(F_1\) on the synthetic CTB-derived benchmark under increasing character-level noise.}
\label{fig:synthetic-results}
\end{figure}

The synthetic benchmark evaluates robustness under controlled character-level perturbation.
For each noise level from 1\% to 10\%, we compare direct segmentation, identical-character projection, and the full two step projection method.
Figure~\ref{fig:synthetic-results} reports token-level \(F_1\) across noise levels.

The synthetic benchmark isolates the effect of increasing source--target divergence.
Direct segmentation degrades as character-level perturbations disrupt the lexical cues used by the segmenter.
Identical-character projection remains effective when most characters are preserved, but its advantage decreases as unresolved substitutions, deletions, and insertions become more frequent.
The full two step projection is more stable across noise levels because it supplements exact character anchors with similarity-based links for residual source--target divergences.

\section{Analysis and discussion}
\label{sec:analysis-discussion}

Chinese word boundary recovery is not ordinary segmentation over a noisy string.
The corrected target supplies lexical boundary structure that the source alone may obscure, but this structure becomes usable only through reliable character correspondence.
The task is therefore governed by two constraints: preserving the learner string and recovering target-licensed lexical units.

The two-step alignment procedure separates preservation from residual correspondence.
\(\mathcal{P}_{1}\) fixes identical characters as high-confidence anchors, and \(\mathcal{P}_{2}\) aligns only the unmatched residual characters.
This residualization changes the effective alignment space: characters separated by anchored material in the original strings may become adjacent after the anchors are removed.
The method therefore preserves a linear projection pipeline while allowing a restricted class of non-local correspondences.
\(\mathcal{P}_{2}\) is not a fully non-monotone alignment model, but a residual alignment model whose flexibility is conditioned on an identity backbone.

\subsection{Residual alignment}
\label{subsec:analysis-two-step}

The gain from \(\mathcal{P}_{1}\) to \(\mathcal{P}_{2}\) reflects the limits of exact character preservation.
Identical characters provide high-precision anchors for shared material, but they do not resolve substitutions, additions, omissions, or locally displaced material.
When such material occurs inside or near a lexicalized target word, \(\mathcal{P}_{1}\) may retain broad source--target coverage without recovering the relevant source-side token span.

\(\mathcal{P}_{2}\) confines similarity-based matching to the residual region left unresolved by \(\mathcal{P}_{1}\), allowing non-identical characters to be linked only when they are visually, phonologically, positionally, or contextually plausible.
In the learner benchmark, alignment coverage increases from \(93.09\%\) to \(95.61\%\) on the source side and from \(91.36\%\) to \(93.84\%\) on the target side.
This improvement suggests that residual character correspondence provides useful evidence for boundary projection beyond exact character preservation.

Although \(\mathcal{P}_{2}\) is embedded in a linear projection pipeline, it can identify a limited class of word-order variation because it operates only on residual characters after exact matching.
Exact anchors established by \(\mathcal{P}_{1}\) are fixed, while the residual step may add links between remaining source and target characters even when their local order differs.
Among the \(1{,}230\) additional source-side character alignments introduced by \(\mathcal{P}_{2}\), \(737\) (\(59.92\%\)) are non-monotone and are therefore counted as word-order variation alignments across 334 sentences.
These links represent only \(1.51\%\) of all source characters and \(1.58\%\) of \(\mathcal{P}_{2}\)-aligned source characters, but they show that residual alignment captures a small and systematic portion of non-monotone learner divergence without requiring a globally non-monotone alignment model.

\subsection{Ablation of similarity features}
\label{subsec:feature-ablation}

\begin{table}[!ht]
\footnotesize
\centering
\begin{tabular}{lc}
\toprule
Method & Token \(F_1\) \\
\midrule
\(\mathcal{P}_{2}\) 
& \textbf{0.9916} \\
\quad w/o glyph similarity
& 0.9899 \\
\quad w/o phonological similarity
& 0.9916 \\
\quad w/o positional similarity
& 0.9899 \\
\quad w/o semantic similarity
& 0.9902 \\
\bottomrule
\end{tabular}
\caption{Ablation of similarity features in the second alignment step on the learner Chinese benchmark.}
\label{tab:feature-ablation}
\end{table}

The ablation results show that glyph and positional similarity provide the main residual evidence.
Removing either feature reduces performance to the level of \(\mathcal{P}_{1}\), indicating that visually related substitutions and local positional bias are both needed after identity anchors have constrained the search space.
Removing semantic similarity yields a smaller drop, while removing phonological similarity does not affect aggregate token-level \(F_1\).
This pattern is consistent with the learner setting: MuCGEC consists of essays by Chinese-as-a-Second-Language learners \citep{zhang-etal-2022-mucgec}, and prior work reports that phonological errors are relatively infrequent in L2 Chinese character production compared with errors involving orthographic and semantic relations \citep{zhang-xing-2023-interaction}.\footnote{By contrast, broader analyses of erroneous Chinese characters across writing and typing report that around 76\% of errors involve phonologically similar characters \citep{liu-etal-2011-visually}, indicating that the relative weight of phonology depends on population and modality.}
The advantage of the full model therefore comes not only from adding features, but from applying them only after high-confidence identity links have fixed the alignment backbone.

\subsection{Comparison with IBM-style alignment}
\label{subsec:analysis-ibm}

IBM-style alignment provides a diagnostic comparison with a corpus-driven, non-monotone alignment model widely used in statistical machine translation \citep{brown-etal-1991-statistical}.
Unlike \(\mathcal{P}_{1}\), it does not require character identity and can induce correspondences across local order differences.
In the present setting, however, its alignments are estimated from distributional evidence over a relatively small learner benchmark, so identical and near-identical character matches dominate, while rare non-identical correspondences receive limited support.
IBM-style projection therefore approximates much of the exact-match behavior of \(\mathcal{P}_{1}\), but adds only those residual correspondences that are sufficiently frequent in the corpus.
On the learner benchmark, it obtains a token-level \(F_1\) of 0.9489, below both \(\mathcal{P}_{1}\) and \(\mathcal{P}_{2}\).
Details of the IBM-style setting are provided in Appendix~\ref{app:ibm-alignment}.

The result reflects a mismatch between corpus-level alignment and learner-character divergence.
The benchmark contains only around one thousand sentence pairs, limiting reliable estimation for sparse non-identical character pairs.
Many learner substitutions are instead motivated by character-internal relations: two characters may be visually similar, phonologically close, or contextually substitutable even when their pairwise co-occurrence is rare.
An IBM-style model can learn frequent correspondences from observed distributions, but it does not encode why an infrequent non-identical pair should be aligned.

The contrast is therefore not monotone versus non-monotone alignment alone.
IBM-style alignment is formally flexible, but its flexibility is corpus-driven.
The proposed residual aligner is more constrained but more targeted: after exact anchors have been fixed, it can recover limited word-order variation through residual links while using character-level similarity features to support rare non-identical correspondences.
For Chinese learner text, the central difficulty is not arbitrary reordering, but the recovery of plausible character correspondences under sparse supervision.

\subsection{Error types}
\label{subsec:error-types}

Remaining tokenization errors are classified as over-segmentation, under-segmentation, or boundary drift.
Over-segmentation splits a gold token into multiple predicted tokens.
Under-segmentation merges multiple gold tokens into one predicted token.
Boundary drift preserves the number of tokens but shifts an internal boundary.

\begin{table}[!ht]
\footnotesize
\centering
\begin{tabular}{lccc}
\toprule
Method & Over-seg. & Under-seg. & Drift \\
\midrule
\(\mathcal{D}\)  & 172 / 75.8\% & 52 / 22.9\% & 3 / 1.3\% \\
\(\mathcal{P}_1\) & 120 / 56.9\% & 88 / 41.7\% & 3 / 1.4\% \\
\(\mathcal{P}_2\) & 88 / 50.0\% & 84 / 47.7\% & 4 / 2.3\% \\
\bottomrule
\end{tabular}
\caption{Distribution of tokenization error types on the learner Chinese benchmark.}
\label{tab:error-type-distribution}
\end{table}

Direct segmentation is dominated by over-segmentation, as learner substitutions often obscure lexicalized compounds and lead the segmenter to split forms whose corrected targets correspond to single compound-level words.
Projection reduces this fragmentation because target-side compounds can license source-side spans.

The transition from \(\mathcal{D}\) to \(\mathcal{P}_{1}\) reduces over-segmentation but increases the relative share of under-segmentation, reflecting the conservativeness of exact-anchor projection.
When residual material cannot be aligned, the operator may preserve or merge spans rather than insert unsupported internal boundaries.
\(\mathcal{P}_{2}\) further reduces over-segmentation, showing that residual links help recover compound-level units obscured by non-identical characters.
Boundary drift remains rare, indicating that most remaining failures concern whether a span should be split or merged, rather than slight boundary displacement.

\subsection{Robustness under increasing noise}
\label{subsec:analysis-robustness}

The synthetic benchmark isolates source--target divergence under controlled noise.
As the noise ratio increases, direct segmentation becomes less reliable because the noisy source moves farther from standard lexical forms.
Projection remains effective only insofar as character alignment can still connect the noisy source to the clean target segmentation.
Thus, \(\mathcal{P}_{1}\) weakens as substitutions, insertions, and deletions enlarge the residual regions left unresolved by exact matching, whereas \(\mathcal{P}_{2}\) degrades more gradually because its second step supplies similarity-based links for part of this residual material.
The synthetic results therefore show that the learner-benchmark gains reflect a broader robustness advantage under controlled source--target divergence.

\section{Conclusion}

This paper introduced a two-step character alignment projection framework for Chinese word boundary recovery in learner text.
The task is framed as target-conditioned projection rather than ordinary segmentation over a noisy string: the corrected target supplies lexical boundary structure, while the learner source determines the string that must be preserved.
The framework first constructs an identical-character alignment backbone, then resolves residual correspondences with similarity-based evidence, and finally transfers target-side boundaries through a deterministic projection operator.

Experiments on a manually checked MuCGEC-based learner benchmark and a synthetic CTB-derived benchmark show that target-conditioned projection improves over direct segmentation, and that residual alignment improves over projection based only on identical-character anchors.
The analysis shows that \(\mathcal{P}_{2}\) increases alignment coverage, recovers a limited but systematic class of word-order variation, and remains more robust under controlled source--target divergence.
The procedure is granularity-parametric: the present experiments use CTB-style compound-level segmentation, but the same operator can project boundary conventions supplied by any segmented target.

\section*{Limitations}

The proposed method assumes that the corrected target provides a useful structural guide for the learner source.
This assumption is appropriate for grammatical error correction data, but it may weaken when the correction substantially rewrites the source.
Large paraphrases, extensive reordering, or corrections that introduce new content can create alignment regions where neither identical-character anchoring nor similarity-based matching is reliable.

The method also depends on the quality and convention of the target-side segmentation.
Projection can only transfer the boundaries supplied to it.
If the target segmentation is inconsistent or follows a different convention from the evaluation reference, the recovered source tokenization will inherit these inconsistencies.
The present benchmarks use compound-based segmentation, so the reported results should be interpreted under that granularity rather than as convention-independent word segmentation accuracy.

Finally, the similarity-based second step introduces feature weights and thresholds that must be fixed or tuned.
Although the projection operator is deterministic for a given alignment, alignment quality depends on these design choices.
The phonological similarity component is also coarse, since the current pinyin similarity treats tone marks as ordinary characters.
As a result, tone-only contrasts such as \zh{妈} \textit{mā} and \zh{麻} \textit{má} may be scored comparably to pairs with broader segmental differences such as \zh{妈} \textit{mā} and \zh{八} \textit{bā}, which may partly explain why phonological similarity contributes little in the present ablation.
Future work should examine whether these parameters and similarity definitions transfer across learner corpora, segmentation conventions, and other types of non-standard Chinese text.


\appendix

\section{Background}
\label{app:background}

Chinese word segmentation has long been treated as a foundational preprocessing task.
The SIGHAN Bakeoff datasets \citep{sproat-emerson-2003-first} established shared benchmarks for segmentation, while resources such as the Chinese Treebank \citep{xue-etal-2005-ctb}, the PKU corpora \citep{shiwen-etal-2002-pekin}, and Universal Dependencies for Chinese \citep{nivre-etal-2020-universal} provide segmentation conventions tied to broader linguistic annotation.
These resources are essential for evaluating segmentation under fixed corpus-specific guidelines.
They do not, however, directly address the setting considered in this paper: recovering source-side word boundaries for a noisy learner sentence by using the segmented correction as a structural guide.

This setting arises naturally in Chinese grammatical error correction.
A learner sentence may contain substitutions, omissions, additions, or local word-order changes, while the corrected sentence provides a cleaner version of the intended expression.
If the learner source and corrected target are segmented independently, the resulting word boundaries may become incompatible.
Such incompatibilities affect edit alignment, error classification, and evaluation, because edits are then computed over different token units.
The relevant task is therefore not ordinary segmentation of a single string, but word boundary recovery over a paired source--target instance.
The recovered source segmentation must cover the original learner sentence while remaining consistent with the target-side segmentation wherever the two sentences correspond.

The proposed method follows the general logic of annotation projection, where linguistic structure is transferred through alignment \citep{yarowsky-ngai-2001-inducing,hwa-EtAl:2005}.
Unlike cross-lingual projection, the present setting is monolingual and character-based.
Chinese characters provide a stable substrate below the word level, while word boundaries are the structures to be recovered.
Because learner input may involve non-identical but related characters, exact matching alone is insufficient.
The proposed two step alignment strategy therefore first fixes identical characters as monotone high-confidence anchors, and then aligns unresolved characters using glyph, phonological, positional, and contextual semantic similarity.

\section{Similarity features for residual alignment}
\label{app:similarity-features}

This appendix specifies the residual alignment model used in \(\mathcal{P}_2\).
Residual alignment is applied only after the exact-match alignment in \(\mathcal{P}_1\).
Let \(U_S\) and \(U_T\) denote the source and target character indices that remain unaligned after \(\mathcal{P}_1\).
For each residual target character \(t_j\), the model searches for a source character \(s_i\), where \(i\in U_S\) and \(j\in U_T\), using four similarity features.

\paragraph{Glyph similarity}

Glyph similarity captures visual resemblance between Chinese characters.
We use the precomputed character-pair similarity table of \citep{gu-etal-2025-improving}.
The table is loaded as a set of bidirectional character-pair scores, so that a pair \((a,b)\) and its reverse \((b,a)\) receive the same value.
For a candidate pair \((s_i,t_j)\), the glyph feature is
\[
\mathrm{sim}_{\mathrm{glyph}}(s_i,t_j)\in[0,1].
\]
Pairs absent from the table receive score 0 and are therefore treated as having no glyph-based similarity.
This feature targets substitutions involving visually related characters.

\paragraph{Phonological similarity}

Phonological similarity captures pronunciation-based substitutions.
Each character is converted to pinyin with tone information using the Python library \texttt{pypinyin}.
Similarity is computed from normalized Levenshtein distance:
\[
\mathrm{sim}_{\mathrm{pinyin}}(s_i,t_j)
=
1-
\frac{
\mathrm{Lev}(\mathrm{py}(s_i),\mathrm{py}(t_j))
}{
\max(|\mathrm{py}(s_i)|,|\mathrm{py}(t_j)|)
}.
\]
Here, \(\mathrm{py}(\cdot)\) denotes the pinyin representation of a character, and \(\mathrm{Lev}(\cdot,\cdot)\) denotes Levenshtein distance.
Tone is encoded as part of the pinyin string.
Thus, tone-only differences are penalized in the same way as other single-character pinyin differences.
This design keeps the implementation simple, but it may understate phonological relatedness when two characters differ only in tone.

\paragraph{Positional similarity}

Positional similarity provides a weak locality preference.
For source length \(m\) and target length \(n\), the implementation defines
\[
\mathrm{sim}_{\mathrm{pos}}(s_i,t_j)
=
1-
\frac{|i-j|}{\max(m,n)}.
\]
This feature does not impose monotonicity.
It only favors residual links whose source and target indices are close.

\paragraph{Embedding similarity}

Embedding similarity measures contextual compatibility between source and target characters.
We encode the source and target character sequences with \texttt{bert-base-chinese}, using the HuggingFace \texttt{transformers} implementation.
Let \(\mathbf{h}_{s_i}\) and \(\mathbf{h}_{t_j}\) be the final-layer hidden states corresponding to \(s_i\) and \(t_j\).
The embedding feature is cosine similarity:
\[
\mathrm{sim}_{\mathrm{emb}}(s_i,t_j)
=
\frac{
\mathbf{h}_{s_i}\cdot \mathbf{h}_{t_j}
}{
\|\mathbf{h}_{s_i}\|\,\|\mathbf{h}_{t_j}\|
}.
\]
The implementation uses this cosine value directly in the interpolated score.

\paragraph{Feature interpolation}

The residual alignment score is a weighted interpolation of the four features:
\[
\begin{aligned}
\mathrm{score}(s_i,t_j)
&=
\lambda_{\mathrm{emb}}\mathrm{sim}_{\mathrm{emb}}(s_i,t_j) \\
&\quad+
\lambda_{\mathrm{glyph}}\mathrm{sim}_{\mathrm{glyph}}(s_i,t_j) \\
&\quad+
\lambda_{\mathrm{pinyin}}\mathrm{sim}_{\mathrm{pinyin}}(s_i,t_j) \\
&\quad+
\lambda_{\mathrm{pos}}\mathrm{sim}_{\mathrm{pos}}(s_i,t_j).
\end{aligned}
\]
The submitted system uses the following setting:
$\lambda_{\mathrm{emb}}=0.1, 
\lambda_{\mathrm{glyph}}=0.6, 
\lambda_{\mathrm{pinyin}}=0.0, 
\lambda_{\mathrm{pos}}=0.3, 
\tau=0.85$.
Although the phonological feature is implemented, its selected weight is 0 in the final configuration.

\paragraph{Threshold and weight selection}

Weights and threshold are selected on the development data using a grid-search procedure.
We perform 5-fold cross-validation.
For each candidate weight vector and threshold, the full projection pipeline is run and the resulting segmentation is evaluated against the gold source segmentation using token-level \(F_1\).
The candidate setting is ranked by mean cross-validation token-level \(F_1\), with the maximum fold score and the score distribution used as additional diagnostics.
The default search uses weight increments of 0.1.
Thresholds are searched from 0.30 to 0.90 in increments of 0.05.
The selected setting is then fixed and applied to the evaluation set.
No test-set labels are used for selecting either the weights or the threshold.

\begin{figure}[!ht]
\centering
\includegraphics[width=\linewidth]{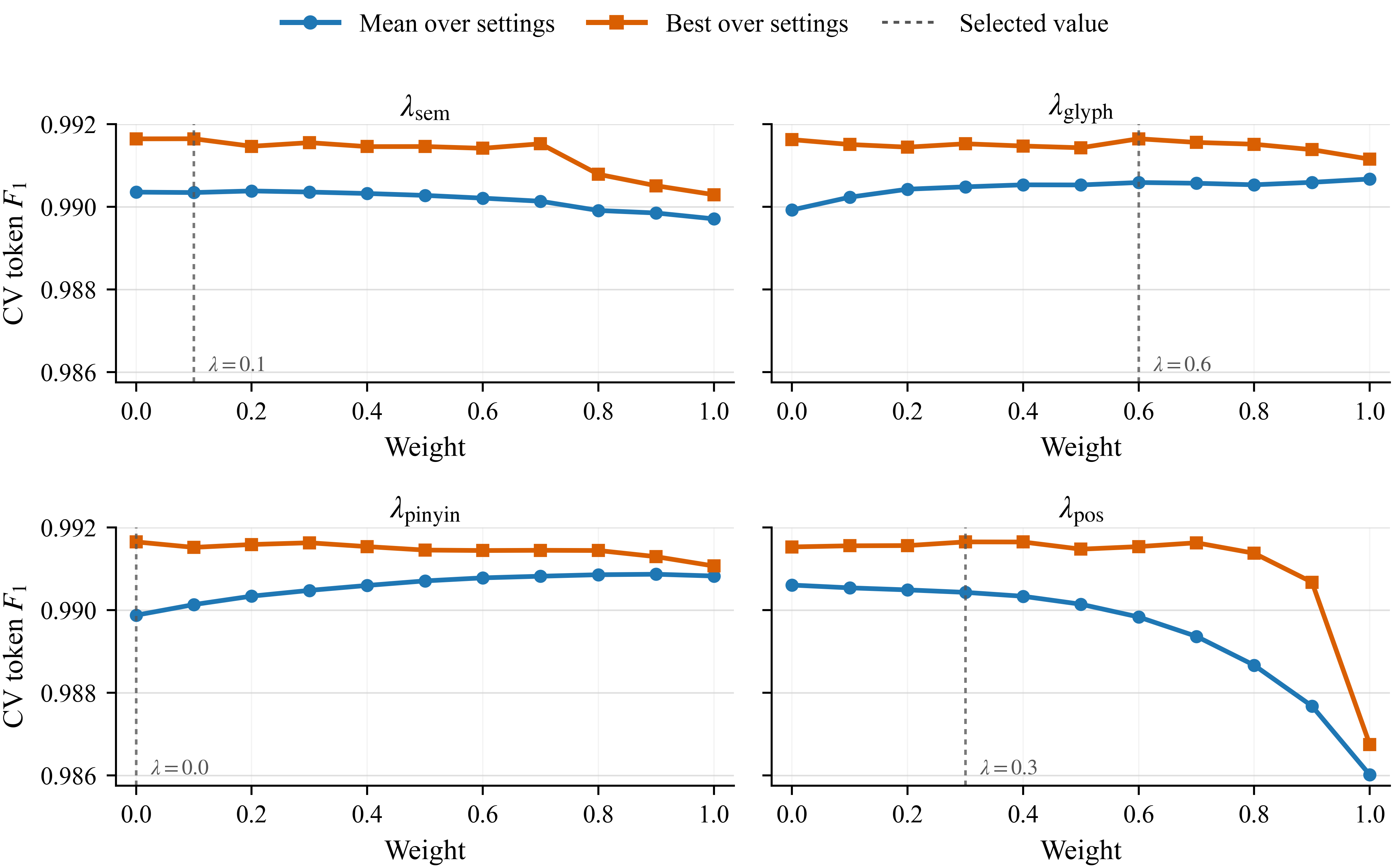}
\caption{Cross-validation sensitivity to individual \(\mathcal{P}_2\) feature weights.}
\label{fig:step2-weight-sensitivity}
\end{figure}

Figure~\ref{fig:step2-weight-sensitivity} summarizes the sensitivity of token-level \(F_1\) to individual feature weights in the cross-validation search.
Glyph similarity gives the strongest and most stable contribution, while positional similarity helps only as a secondary feature.
Large positional weights sharply reduce performance, showing that locality alone is insufficient for residual character alignment.
Embedding and pinyin weights have smaller effects under this grid, supporting the final configuration with dominant glyph weight, secondary positional weight, minor embedding weight, and zero pinyin weight.

\paragraph{Residual alignment selection}

Residual alignment is target-to-source.
For each unaligned target character \(t_j\), considered in target order, the model searches over unaligned source characters that have not already been assigned by \(\mathcal{P}_2\).
The highest-scoring source character is selected only if its score is greater than \(\tau\).
Once a source character has been selected, it cannot be reused for another residual target character.
This produces a one-to-one residual alignment from a subset of \(U_T\) to a subset of \(U_S\).
Characters that cannot be linked under this condition remain unresolved.

\paragraph{Projection after residual alignment}

The residual alignment model does not directly insert or delete word boundaries.
It only supplies additional character links to the same deterministic projection operator used after \(\mathcal{P}_1\).
The initial source segmentation is obtained with the automatic word segmentation tool.
Target word boundaries are projected only when adjacent target characters align to adjacent source characters.
Internal source boundaries are removed only when all characters of a multi-character target token align to a contiguous source span and the target token length matches the aligned source span length.
This span-length constraint prevents the projection operator from merging across extra source characters that happen to occur between aligned endpoints.
Thus, \(\mathcal{P}_2\) can improve alignment coverage for non-identical characters, but all boundary changes remain subject to the same conservative projection constraints.

\section{Formal properties and complexity of projection}
\label{app:projection-properties}

We state basic properties of the projection operator \(\Pi\) for any fixed character alignment \(\mathcal{A}\).
These properties depend only on the alignment relation supplied to Algorithm~\ref{alg:projection}, not on the method used to obtain it.
The same statements therefore apply to both projection settings used in this paper: \(\mathcal{P}_1\), which uses identical-character correspondences, and \(\mathcal{P}_2\), which augments these correspondences with similarity-based residual links.

\begin{lemma}
Given \((S,W_S^{0},W_T,\mathcal{A})\), Algorithm~\ref{alg:projection} returns a projected source segmentation \(\widehat{W}_S\) such that:
(i) every source character \(s_i\) is covered by exactly one token in \(\widehat{W}_S\);
(ii) every token in \(\widehat{W}_S\) is a contiguous span of \(S\);
(iii) every inserted boundary is licensed by adjacent target-to-source character correspondence;
(iv) every removed boundary is internal to a contiguous source span licensed by a target token; and
(v) \(\widehat{W}_S\) is a deterministic function of \(S\), \(W_S^{0}\), \(W_T\), and \(\mathcal{A}\).
\end{lemma}

\begin{proofsketch}
The algorithm represents a source segmentation as a set \(B_S\) of word-final source character positions.
Since \(B_S\) is converted back into a segmentation by inserting word breaks into the original source string \(S\), the resulting \(\widehat{W}_S\) necessarily partitions \(S\) into non-overlapping contiguous spans.
Thus, every source character is covered exactly once, and no emitted token can be discontinuous.

Boundary insertion is restricted to target boundaries whose adjacent target characters align to adjacent source characters.
Hence, every inserted boundary is licensed by a local target-to-source character correspondence.
Boundary removal is restricted to source-internal boundaries inside the projected span of a target token.
Such a removal is performed only when the target token projects to a complete contiguous source span.
All other cases are skipped, leaving the current source boundary set unchanged.

Because the initialization from \(W_S^{0}\), the extraction of \(B_T\), and all insertion and removal tests are fixed by \(S\), \(W_S^{0}\), \(W_T\), and \(\mathcal{A}\), the output segmentation is deterministic.
\end{proofsketch}

\paragraph{Complexity}
The projection operator assumes that a character alignment \(\mathcal{A}\) is already given.
Under this assumption, projection is linear in the size of the source string, target string, and alignment representation.
Initializing the source boundary set from \(W_S^{0}\) and extracting the target boundary set from \(W_T\) take \(O(m+n)\) time, where \(m=|S|\) and \(n=|T|\).
Projecting target boundaries requires constant-time alignment lookup for the adjacent target characters around each boundary, and therefore takes \(O(|B_T|)\) time.
Projecting target tokens as source spans requires a single pass over target characters and takes \(O(n)\) time, assuming constant-time access to \(\mathcal{A}\) and constant-time membership queries in \(B_S\).
The final conversion from \(B_S\) to \(\widehat{W}_S\) takes \(O(m)\) time.
Thus, for a fixed alignment, Algorithm~\ref{alg:projection} runs in
\[
O(m+n+|\mathcal{A}|)
\]
time and uses \(O(m)\) auxiliary space for the source boundary set.

The cost of constructing \(\mathcal{A}\) is external to the projection algorithm.
In our pipeline, Step~1 obtains identical-character correspondences, while Step~2 scores unresolved character pairs using similarity-based evidence.
These alignment steps differ in how they produce character correspondences, but once \(\mathcal{A}\) is fixed, the projection operator \(\Pi\) is unchanged.

\section{Benchmark construction details}
\label{app:benchmark-details}

This appendix provides supplementary details on the construction of the two benchmarks introduced in Section~\ref{sec:benchmarks}.
The main text describes the reference selection procedure, segmentation granularity, source-side annotation protocol, and synthetic noise design.
Here we report additional annotation statistics and clarify the token-level evaluation used in the experiments.
Table~\ref{tab:benchmarks} summarizes the two evaluation settings.

\begin{table*}[!t]
\footnotesize
\centering
\begin{tabular}{lccc}
\toprule
Benchmark & Source & Pairs & Purpose \\
\midrule
Learner & MuCGEC dev & 1,079 & Natural learner errors \\
Synthetic & CTB 5.1 & 1,079 $\times$ 10 & Controlled character-level noise \\
\bottomrule
\end{tabular}
\caption{Evaluation benchmarks for Chinese word boundary recovery.}
\label{tab:benchmarks}
\end{table*}

\subsection{Reference selection outcomes}
\label{app:reference-selection}

Table~\ref{tab:ref-selection} summarizes the manual reference selection outcomes for the 314 MuCGEC instances that were not resolved automatically.
Three annotators independently selected the most appropriate corrected target for each learner source, using grammatical well-formedness and semantic fidelity as the two decision criteria.
Final targets were determined by majority vote.

\begin{table}[!ht]
\footnotesize
\centering
\begin{tabular}{lcccc}
\toprule
& Full & Two-of-three & Disputed & Total \\
\midrule
Sentences & 115 & 199 & 0 & 314 \\
Percentage & 36.6 & 63.4 & 0.0 & 100.0 \\
\bottomrule
\end{tabular}
\caption{Inter-annotator outcomes in MuCGEC reference selection.}
\label{tab:ref-selection}
\end{table}

The full-agreement rate among manually adjudicated cases is 36.6\%, and the average pairwise agreement is 57.7\%.
This level of agreement is expected in a setting where several corrected targets may be acceptable.
Since word boundary recovery requires one corrected target for projection, majority voting is used to obtain a deterministic source--target pair.

\subsection{Learner benchmark statistics}
\label{app:learner-statistics}

Table~\ref{tab:learner-statistics} reports the size of the manually annotated learner benchmark.
Source and target token counts are computed after applying the compound-based segmentation guideline described in Section~\ref{subsec:benchmark-learner}.

\begin{table}[!ht]
\footnotesize
\centering
\begin{tabular}{lccc}
\toprule
 & Sentence & Source & Target \\
Dataset & pairs & tokens & tokens \\
\midrule
MuCGEC & 1,079 & 32,536 & 33,140 \\
\bottomrule
\end{tabular}
\caption{Statistics of the manually annotated learner benchmark.}
\label{tab:learner-statistics}
\end{table}

The source-side segmentation was produced by manually correcting system-generated boundary proposals.
Three annotators independently revised the proposals, and residual disagreements were adjudicated.
Boundary-level pairwise \(F_1\) averages 0.987 across annotator pairs, and Krippendorff's \(\alpha\) reaches 0.981, indicating that the compound-based guideline yields highly consistent source-side word boundary annotation.

\subsection{Synthetic benchmark statistics}
\label{app:synthetic-statistics}

The synthetic benchmark is constructed from a fixed CTB 5.1 subset, using the same clean target sentences across all noise levels.
For each noise ratio \(r\in\{1\%,2\%,\ldots,10\%\}\), character-level substitutions, deletions, and insertions are sampled according to the learner-derived distribution:
$p(\text{substitution}) = 0.483$,
$p(\text{deletion}) = 0.318$, and
$p(\text{insertion}) = 0.199$.
The clean CTB sentence remains the target, while the perturbed sentence becomes the noisy source.
Inserted source characters have no corresponding characters in the clean target, while deleted target characters are absent from the noisy source sequence. 
This setup yields ten controlled evaluation conditions, allowing robustness to be measured as source--target divergence increases.

\subsection{Evaluation granularity}
\label{app:evaluation-granularity}

Both benchmarks are evaluated under the same compound-based segmentation granularity.
System outputs are scored at the token level.
A predicted token is counted as correct only when its character span exactly matches a gold source-side token span.
This evaluation directly reflects the goal of word boundary recovery: the system must recover complete word units on the noisy source side, not merely identify isolated boundary offsets.

For the learner benchmark, gold token spans are obtained from the manually adjudicated source-side segmentation.
For the synthetic benchmark, gold token spans are derived from the CTB segmentation after applying the same character-level perturbations used to construct the noisy source.
In both cases, we report micro-averaged token-level \(F_1\) over the full benchmark.

\section{IBM-style character alignment}
\label{app:ibm-alignment}

We use an IBM-style alignment model as a diagnostic comparison for the proposed two step aligner.
The model is not part of the main projection framework; it provides an alternative source of character correspondences \(\mathcal{A}\) for the same projection operator.

Let \(S=s_0\cdots s_{m-1}\) be the source string, and let the segmented target sentence be \(W_T=w^T_1,\ldots,w^T_K\).
Each source character \(s_j\), for \(j\in\{0,\ldots,m-1\}\), is linked to a latent target token index
\[
a_j\in\{0,\ldots,K\},
\]
where \(a_j=0\) denotes the \textsc{null} token.
Under an IBM Model~2-style parameterization, the alignment probability for \(s_j\) factorizes into a distortion term and a token-conditioned character emission term:
\[
\begin{aligned}
p(a_j=i, s_j \mid j,m,K,W_T) \qquad \qquad \\
\qquad \qquad = d(j\mid i,m,K)\,
p(s_j\mid w^T_i),
\end{aligned}
\]
where \(d(j\mid i,m,K)\) models positional distortion and \(p(s_j\mid w^T_i)\) models the probability of generating the source character from the aligned target token.
The marginal likelihood of a source sentence is
\[
\prod_{j=0}^{m-1}
\sum_{i=0}^{K}
d(j\mid i,m,K)\,
p(s_j\mid w^T_i).
\]
Parameters are estimated by expectation--maximization over the sentence pairs.
At decoding time, each source character is assigned its highest-probability target index:
\[
\hat{a}_j
=
\arg\max_{i\in\{0,\ldots,K\}}
d(j\mid i,m,K)\,
p(s_j\mid w^T_i).
\]
Decoding runs in \(O(mK)\) time per sentence.

The resulting token-level alignment sequence is converted into a character-level relation set \(\mathcal{A}\) for projection.
If \(s_j\) is aligned to target token \(w_i^T\), we expand \(w_i^T\) to its target character span and create the corresponding source-to-target character links according to the best local character match within that span.
If \(s_j\) is aligned to \textsc{null}, it remains unresolved.
The resulting \(\mathcal{A}\) is then supplied to the same projection operator used for \(\mathcal{P}_{1}\) and \(\mathcal{P}_{2}\).

IBM-style alignment is useful because it can induce non-monotone links and does not require source and target characters to be identical.
Its flexibility, however, is distributional rather than character-internal.
Lexical probabilities are estimated from observed co-occurrence, so the model does not directly encode visual similarity, phonological similarity, or contextual semantic similarity between individual characters.
This makes it a strong diagnostic baseline for testing whether explicit similarity-aware residual alignment provides benefits beyond corpus-level probabilistic alignment.

\section{Why projection fails}
\label{why-projection-fails}

This section analyzes representative cases of under-segmentation, the error type whose relative share increases under projection.
We use \textsc{g} to indicate a segmentation that matches the gold annotation, and \textsc{b} to indicate an incorrect word boundary decision.
Thus, a pattern such as \textsc{g b b} means that direct segmentation is correct, whereas both projected outputs are incorrect.
{These cases are informative because projection can reduce source-side fragmentation while also preserving locally plausible character groupings that the gold annotation splits.}

\paragraph{Example 1: \textsc{b b b}} In~\eqref{Sorry-I-did}, the learner sentence \zh{我没看你的新写字。} \textit{wǒ méi kàn nǐ de xīn xiě zì} (`I didn't look at your newly written characters.') contains the span \zh{新写字} \textit{xīn xiě zì} (`new-write-characters'), which is not a typical standard grammatical form.
The direct segmentation (\(\mathcal{D}\)) produces \zh{新} \textit{xīn} (`new')$\sqcup$\zh{写字} \textit{xiězì} (`write characters'), a plausible surface analysis because \zh{写字} \textit{xiězì} forms a conventional verb-object compound.

\begin{exe}
\ex \label{Sorry-I-did}
\gll \zh{对不起} \zh{，} \zh{我} \zh{没} \zh{看} \zh{你} \zh{的} \zh{新} \zh{写} \zh{字} \zh{。} \\
\textit{duìbuqǐ} \textit{，} \textit{wǒ} \textit{méi} \textit{kàn} \textit{nǐ} \textit{de} \textit{xīn} \textit{xiě} \textit{zì} \textit{。} \\
sorry , 1\textsc{sg} \textsc{neg} see 2\textsc{sg} \textsc{de} new write character . \\
\glt `Sorry, I didn’t look at your newly written characters.' 
\end{exe}

\begin{center}
\footnotesize
\centering
\begin{tabular}{p{0.4\textwidth} c}
{\zh{对不起$\sqcup$，$\sqcup$我$\sqcup$没$\sqcup$看$\sqcup$你$\sqcup$的$\sqcup$新$\sqcup$写$\sqcup$字$\sqcup$。}} & ($\mathcal{G}$)  \\  
{\zh{对不起$\sqcup$，$\sqcup$我$\sqcup$没$\sqcup$看$\sqcup$你$\sqcup$的$\sqcup$新$\sqcup$写字$\sqcup$。}} & ($\mathcal{D}$)\\
{\zh{对不起$\sqcup$，$\sqcup$我$\sqcup$没$\sqcup$看$\sqcup$你$\sqcup$的$\sqcup$新$\sqcup$写字$\sqcup$。}} & ($\mathcal{P}_1$) \\
{\zh{对不起$\sqcup$，$\sqcup$我$\sqcup$没$\sqcup$看$\sqcup$你$\sqcup$的$\sqcup$新$\sqcup$写字$\sqcup$。}} & ($\mathcal{P}_2$)
\end{tabular}
\end{center}

{The gold segmentation splits the span as \zh{新} \textit{xīn} (`new')$\sqcup$\zh{写} \textit{xiě} (`write')$\sqcup$\zh{字} \textit{zì} (`characters'), a compositional analysis supported by the target-side correction \zh{新写的字} \textit{xīn xiě de zì} (`newly written characters').}
{Under this analysis, \zh{新} \textit{xīn} modifies \zh{写} \textit{xiě}, and \zh{字} \textit{zì} is treated as an independent nominal unit.}
The projected outputs preserve the frequent lexical grouping \zh{写字} \textit{xiězì}, so the boundary between \zh{写} \textit{xiě} and \zh{字} \textit{zì} is not introduced.
This yields under-segmentation relative to the gold standard.

\paragraph{Example 2: \textsc{g b b}}
In~\eqref{and-every}, the learner sentence \zh{而每一代的年轻时代都有不同的音乐调。} 
\textit{ér měi yí dài de nián qīng shí dài dōu yǒu bù tóng de yīn yuè diào} 
(`Each generation’s youth period has different musical tones.') 
contains the span \zh{音乐调} \textit{yīn yuè diào} (`music-tone').
{The target-side correction uses \zh{音乐曲调} \textit{yīn yuè qǔ diào} (`musical tune') in the corresponding position, supporting the gold segmentation \zh{音乐} \textit{yīnyuè} (`music')$\sqcup$\zh{调} \textit{diào} (`tone').}
{This analysis treats \zh{调} \textit{diào} as an independent nominal element, with \zh{音乐} \textit{yīnyuè} functioning as its modifier.}

\begin{exe}
\ex \label{and-every}
\gll \zh{而} \zh{每} \zh{一代} \zh{的} \zh{年轻} \zh{时代} \zh{都} \zh{有} \zh{不同} \zh{的} \zh{音乐} \zh{调} \zh{。} \\
\textit{ér} \textit{měi} \textit{yídài} \textit{de} \textit{niánqīng} \textit{shídài} \textit{dōu} \textit{yǒu} \textit{bùtóng} \textit{de} \textit{yīnyuè} \textit{diào} \textit{。} \\
and each generation \textsc{de} young period has different \textsc{de} music tone . \\
\glt `Each generation’s youth period has different musical tones.'
\end{exe}

\begin{center}
\footnotesize
\centering
\begin{tabular}{p{0.4\textwidth} c}
{\zh{而$\sqcup$每$\sqcup$一代$\sqcup$的$\sqcup$年轻$\sqcup$时代$\sqcup$都$\sqcup$有$\sqcup$不同$\sqcup$的$\sqcup$音乐$\sqcup$调$\sqcup$。}} & ($\mathcal{G}$)  \\  
{\zh{而$\sqcup$每$\sqcup$一代$\sqcup$的$\sqcup$年轻$\sqcup$时代$\sqcup$都$\sqcup$有$\sqcup$不同$\sqcup$的$\sqcup$音乐$\sqcup$调$\sqcup$。}} & ($\mathcal{D}$)\\
{\zh{而$\sqcup$每$\sqcup$一代$\sqcup$的$\sqcup$年轻$\sqcup$时代$\sqcup$都$\sqcup$有$\sqcup$不同$\sqcup$的$\sqcup$音乐调$\sqcup$。}} & ($\mathcal{P}_1$) \\
{\zh{而$\sqcup$每$\sqcup$一代$\sqcup$的$\sqcup$年轻$\sqcup$时代$\sqcup$都$\sqcup$有$\sqcup$不同$\sqcup$的$\sqcup$音乐调$\sqcup$。}} & ($\mathcal{P}_2$)
\end{tabular}
\end{center}

The direct segmentation ($\mathcal{D}$) matches the gold output.
In contrast, both projected outputs merge the span as \zh{音乐调} \textit{yīn yuè diào} (`music-tone').
{Projection therefore preserves a contiguous source-side sequence that is locally plausible, but it does not introduce the boundary supported by the gold annotation.}

\paragraph{Example 3: \textsc{g b b}}
In~\eqref{these-songs}, the learner sentence \zh{这些歌曲，在一定程度上，给人予教育的作用。} 
\textit{zhè xiē gē qǔ, zài yí dìng chéng dù shàng, gěi rén yǔ jiào yù de zuò yòng} 
(`These songs, to a certain extent, have an educational effect on people.') 
contains the non-standard span \zh{给人予} \textit{gěi rén yǔ}.
{The target-side correction uses \zh{给人以} \textit{gěi rén yǐ} (`provide people with'), while the gold segmentation splits the learner form as \zh{人} \textit{rén} (`people')$\sqcup$\zh{予} \textit{yǔ} (`give').}
{This segmentation is consistent with the corrected construction without requiring any claim about learner intention.}

\begin{exe}
\ex \label{these-songs}
\gll \zh{这些} \zh{歌曲} \zh{，} \zh{在} \zh{一定} \zh{程度} \zh{上} \zh{，} \zh{给} \zh{人} \zh{予} \zh{教育} \zh{的} \zh{作用} \zh{。} \\
\textit{zhèxiē} \textit{gēqǔ} \textit{，} \textit{zài} \textit{yídìng} \textit{chéngdù} \textit{shàng} \textit{，} \textit{gěi} \textit{rén} \textit{yǔ} \textit{jiàoyù} \textit{de} \textit{zuòyòng} \textit{。} \\
these songs , at a certain degree LOC , give person give education \textsc{de} effect . \\
\glt `These songs, to a certain extent, have an educational effect on people.'
\end{exe}

\begin{center}
\footnotesize
\centering
\begin{tabular}{p{0.4\textwidth} c}
{\zh{这些$\sqcup$歌曲$\sqcup$，$\sqcup$在$\sqcup$一定$\sqcup$程度$\sqcup$上$\sqcup$，$\sqcup$给$\sqcup$人$\sqcup$予$\sqcup$教育$\sqcup$的$\sqcup$作用$\sqcup$。}} & ($\mathcal{G}$)  \\  
{\zh{这些$\sqcup$歌曲$\sqcup$，$\sqcup$在$\sqcup$一定$\sqcup$程度$\sqcup$上$\sqcup$，$\sqcup$给$\sqcup$人$\sqcup$予$\sqcup$教育$\sqcup$的$\sqcup$作用$\sqcup$。}} & ($\mathcal{D}$)\\
{\zh{这些$\sqcup$歌曲$\sqcup$，$\sqcup$在$\sqcup$一定$\sqcup$程度$\sqcup$上$\sqcup$，$\sqcup$给$\sqcup$人予$\sqcup$教育$\sqcup$的$\sqcup$作用$\sqcup$。}} & ($\mathcal{P}_1$) \\
{\zh{这些$\sqcup$歌曲$\sqcup$，$\sqcup$在$\sqcup$一定$\sqcup$程度$\sqcup$上$\sqcup$，$\sqcup$给$\sqcup$人予$\sqcup$教育$\sqcup$的$\sqcup$作用$\sqcup$。}} & ($\mathcal{P}_2$)
\end{tabular}
\end{center}

The direct segmentation ($\mathcal{D}$) matches the gold output.
Both projected outputs instead produce \zh{人予} \textit{rén yǔ}, merging the two characters into a single unit.
{The error shows that projection may preserve a contiguous learner-side sequence even when the gold annotation separates the corresponding elements.}
As a result, the boundary between \zh{人} \textit{rén} and \zh{予} \textit{yǔ} is not introduced, resulting in under-segmentation.

\section{Error analysis visualization}
\label{app:error-visualization}

We develop an interactive error analysis viewer to inspect the learner benchmark at the sentence level: {\url{https://anonymous.4open.science/w/error-analysis-viewer-EB80}}.
The viewer contains all 1,079 learner sentences and displays, for each instance, the learner source sentence, the corrected target sentence, the gold source-side tokenization, and the outputs of the three evaluated systems: direct segmentation (\(\mathcal{D}\)), identical-character projection (\(\mathcal{P}_{1}\)), and two step projection (\(\mathcal{P}_{2}\)).
Tokens are displayed as character spans, with correct predicted spans marked separately from mismatched spans, allowing local segmentation errors to be inspected directly.

The interface is designed to support qualitative error analysis rather than aggregate scoring alone.
Users can search by sentence identifier, source text, target text, or token content.
They can also filter examples by system status, selecting whether \(\mathcal{D}\), \(\mathcal{P}_{1}\), or \(\mathcal{P}_{2}\) is correct or incorrect.
Additional filters isolate error types such as over-segmentation, under-segmentation, and boundary drift, as well as method-level outcome patterns such as \textsc{b g g} or \textsc{g b g}, where \textsc{g} indicates that a system matches the gold tokenization and \textsc{b} indicates that it does not.
The viewer also includes a dedicated filter for cases where the two step method corrects an error made by direct segmentation or identical-character projection.

For each sentence, the viewer presents the gold tokenization and the three system outputs in parallel.
Predicted tokens whose character spans match the gold source-side span are marked as correct, while mismatched spans are highlighted as errors.
When local errors are present, the viewer reports the error type, the affected character span, the gold token, the predicted tokenization, and a short explanation of the mismatch.
This makes it possible to trace aggregate improvements back to concrete local phenomena, such as compound over-segmentation corrected by projection or residual errors caused by unresolved character divergence.

\begin{figure}[!ht]
    \centering
\subfloat[Example 1: \zh{新写字} \textit{xīnxiězì} \label{fig:fail-e1}]{
\includegraphics[width=.4\textwidth]{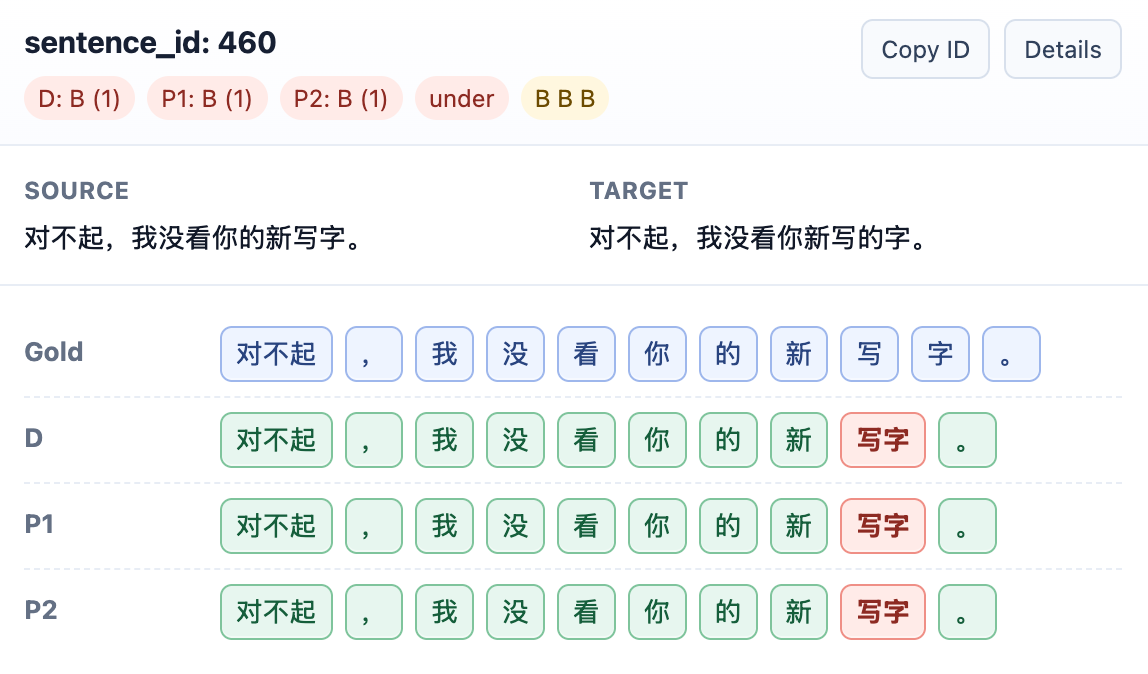}
}

\subfloat[Example 2: \zh{音乐调} \textit{yīnyuèdiào} \label{fig:fail-e2}]{
\includegraphics[width=.45\textwidth]{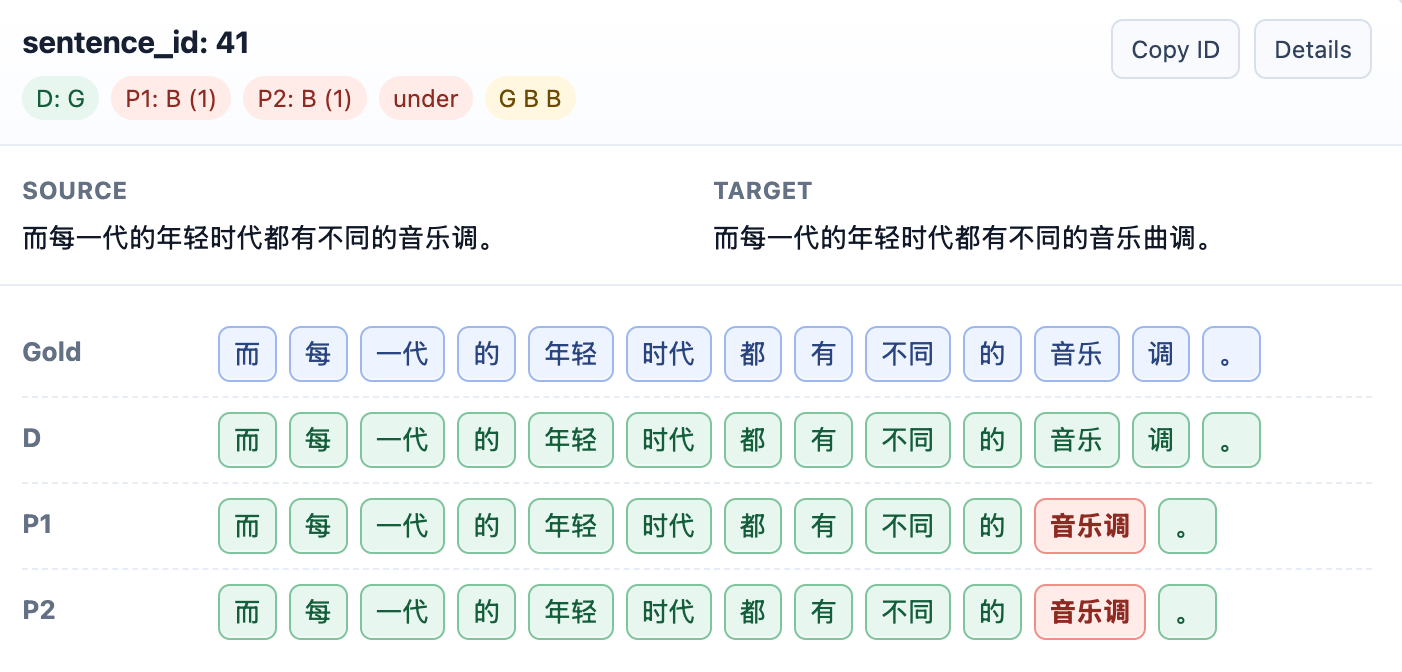}
}

\subfloat[Example 3: \zh{人予} \textit{rényǔ} \label{fig:fail-e3}]{
\includegraphics[width=.45\textwidth]{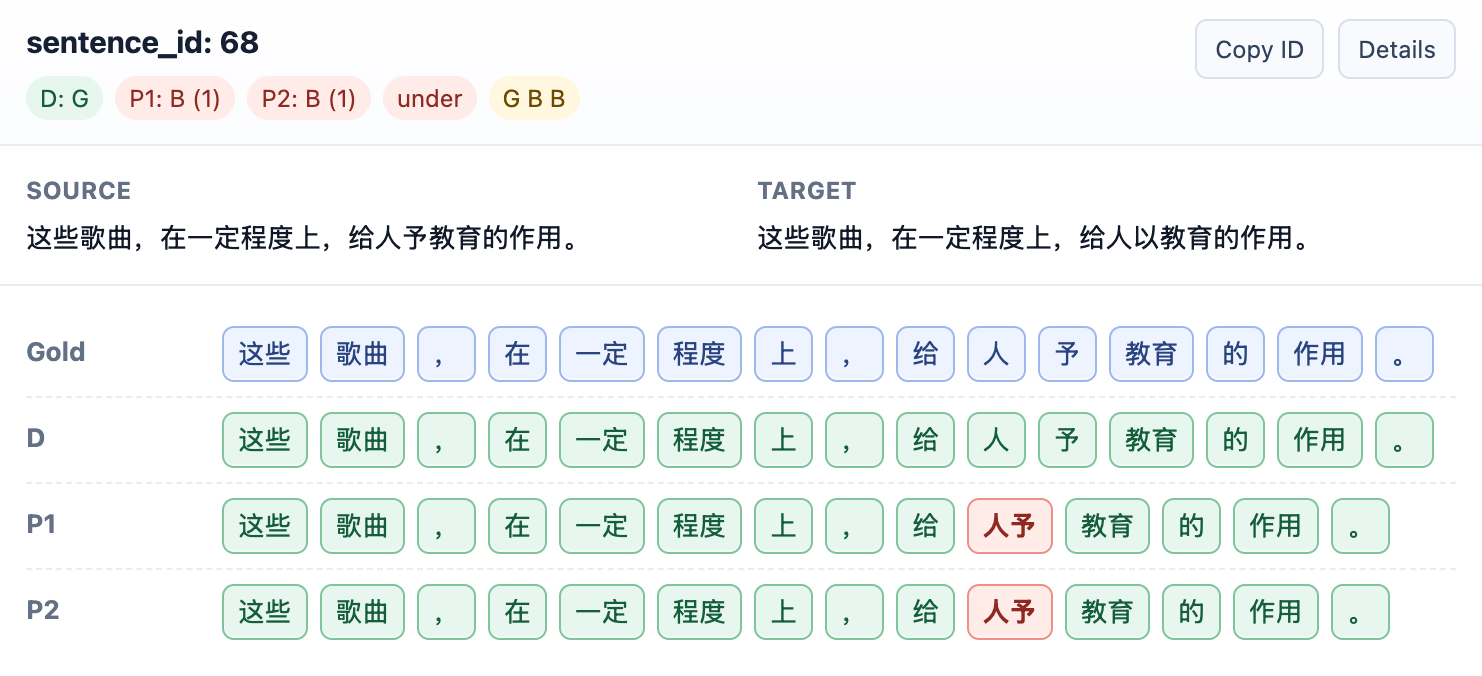}
}
    \caption{Visualization of representative under-segmentation errors discussed in Appendix~\ref{why-projection-fails}. 
    }
    \label{fig:fail-examples}
\end{figure}

The visualization is implemented as a static browser-based tool.
The data file stores the generated 1,079-sentence benchmark output, including source and target strings, gold spans, system tokenizations, error labels, and method-level correctness information.
The frontend renders these instances as searchable and filterable cards, while the accompanying stylesheet defines the color scheme for gold tokens, correct predicted spans, and erroneous predicted spans.
Figure~\ref{fig:fail-examples} visualizes three under-segmentation errors from Appendix~\ref{why-projection-fails}, where projected segmentations fail to introduce gold internal boundaries.

\end{document}